# Improved Automatic Target Recognition in Synthetic Aperture Sonar Imagery Using Large Deep Neural Networks

C.J. Moore, *Student Member, IEEE*, Alex Hurt, *Member, IEEE*, and Jordan Malof., *Member, IEEE*

***Abstract*—Automatic Target Recognition (ATR) in Synthetic Aperture Sonar (SAS) is often underpinned by deep neural networks (DNNs). Recent research has focused on relatively low-capacity DNNs (i.e., relatively fewer trainable parameters) to avoid overfitting on the small training datasets that are typically available for SAS. However, recent research has found that increasing DNN capacity often improves performance, even when training data is scarce. Motivated by this emerging finding - termed the double descent phenomenon - this work investigates the use of especially high-capacity DNNs for SAS-ATR. We evaluate the performance of several state-of-the-art DNN architectures as we increase their model capacity. We examine performance in a realistic leave-one-geo-location-out cross-validation scheme. We find that increasing model capacity often increases performance substantially, and is never harmful. All of our large models significantly outperform two baseline models from recent publications. Training high-capacity models on small dataset however requires careful regularization, and we also perform systematic experiments showing the impact of several regularization strategies: pretraining on other imaging modalities, data augmentation (including some new to SAS), DropPath, and weight decay. We show that each of these strategies independently improves the performance of the large models.**

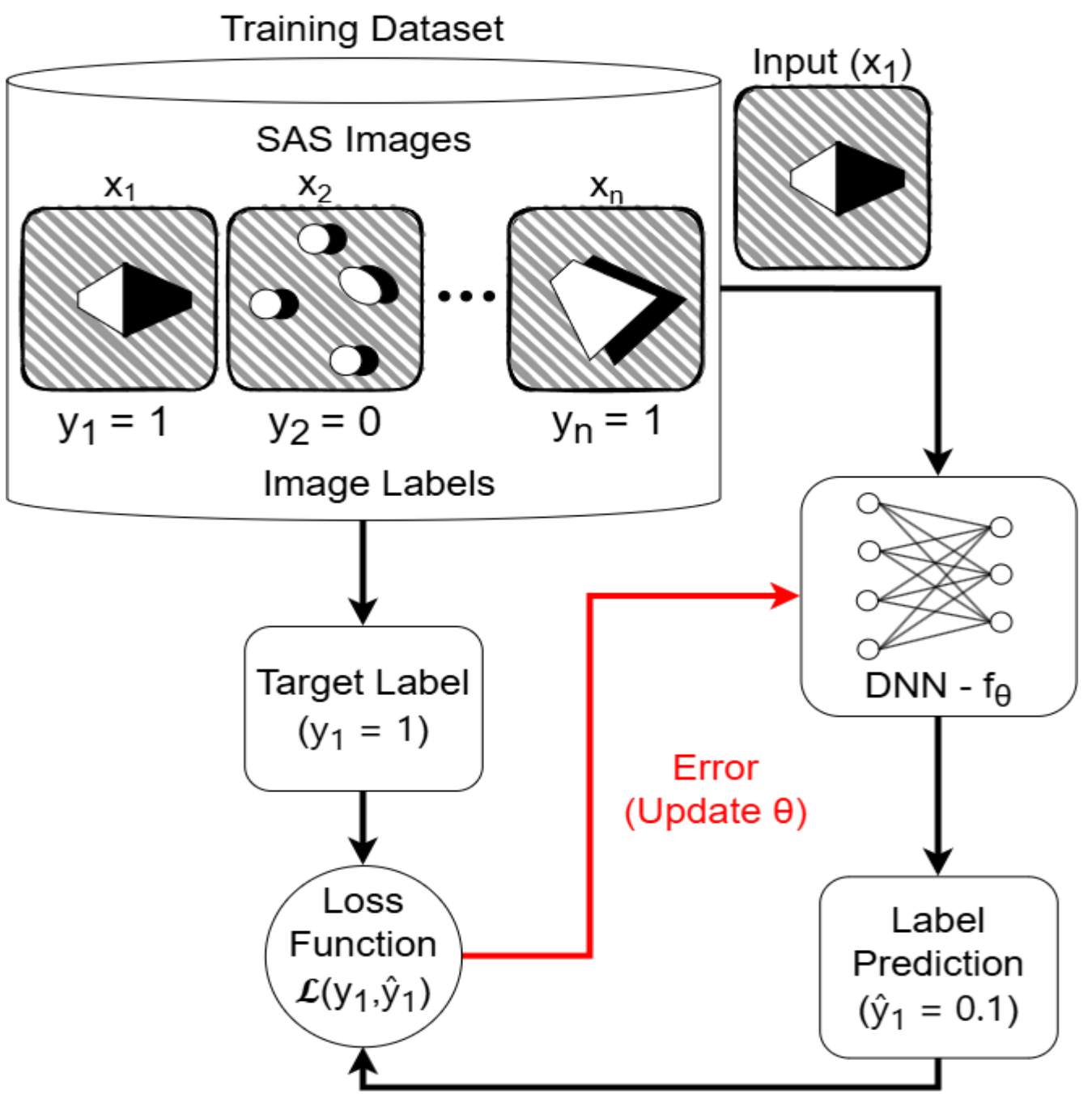

**Fig. 1. Illustration of training deep neural networks (DNNs) for SAS-ATR.**

## I. Introduction

This work focuses on the problem of automatic target recognition (ATR) for synthetic aperture sonar (SAS) data. Many objects of interest can be detected and classified in SAS imagery [1], [2], [3], [4], but SAS imagery of the seafloor is voluminous, making it costly to inspect it manually. SAS-ATR algorithms are designed to efficiently process large volumes of SAS imagery and indicate where target objects are likely to exist. One promising class of SAS-ATR algorithms is based upon deep neural networks (DNNs) [5], [6], [7]. DNNs are a class of mathematical *models* of the form $y = f_\theta(x)$ with adjustable parameters $\theta$ that control the mapping from input, $x$, to output, $y$. In the context of SAS-ATR, $x$ may correspond to a SAS image, and $y$ may correspond to a binary label indicating whether a target object is present. The DNN model is *trained* to produce accurate output using pairs of real-world SAS imagery and their true labels (i.e., target or non-target), termed *training data*. During the training process (**Fig. 1**), the parameters $\theta$ are gradually adjusted so that the DNN makes accurate predictions on the training data. Once trained, DNNs can be deployed to make predictions for novel settings of $x$ that were not present in the training data.

One potential problem with DNNs is that they can learn spurious correlations between the input and output in their training data, causing them to make poor predictions on novel input data, termed poor generalization. This is a widely studied phenomenon in machine learning termed *overfitting*. Until recently, the theory of machine theory predicts that the risk of overfitting increases in proportion to model capacity (i.e., number of free parameters), or as the quantity of training data decreases [8]. Motivated by this theory, most recent SAS-ATR research has explored DNNs with relatively low capacity (i.e., few trainable parameters) (e.g., [7], [9], [10], [11]), due to the tendency for there to be limited quantities of labeled data to train SAS-ATR models [12], [13]. However, recent research on deep learning (often focused on DNNs) has provided both empirical and theoretical evidence that higher-capacity DNNs are often more performant than small DNNs, even when training data is relatively limited [14], [15]. This phenomenon, known as "Double Descent" [16] has shown that while models initially begin to overfit as their capacity increases, their generalization eventually begins to improve again as the number of trainable parameters increases (see **Fig. 2**). Crucially, the most performant networks are often highly overparameterized (i.e. they have many more trainable parameters than training data points). This observation has motivated the use of large networks in many application areas even when training on small datasets [17], [18].

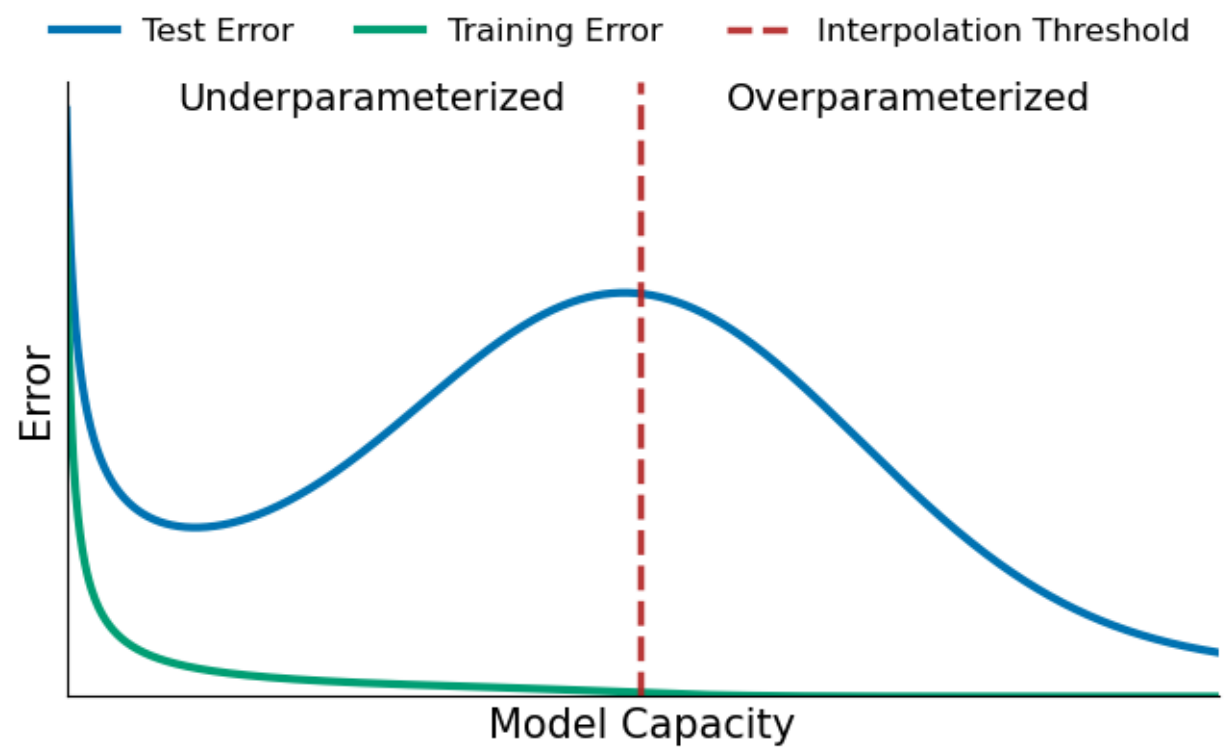

**Fig. 2. Illustration of the Double Descent phenomenon. Model generalization error (e.g., performance on novel input) at first increases as model capacity increases and then reduces once model capacity exceeds the interpolation threshold, where there are more model parameters than can be uniquely determined by the available training data.**

### *A. Contributions of This Work.*

In this work we seek to leverage these recent findings on double descent to improve the overall performance of SAS-ATR**.** The overarching question for this paper is then as follows: can recent large vision models achieve state-of-the-art accuracy on SAS-ATR? To this end, we investigate DNNs for SAS-ATR with substantially higher capacity than those used in previous studies. Motivated by this goal, we also investigate transformer-based DNNs. Transformers are DNN models with fundamentally different connectivity, or architecture, compared to convolutional models, but they also often possess greater capacity than convolutional models as well. Although transformers have received state-of-the-art performance on many vision tasks, they have received relatively limited attention in the SAS-ATR literature thus far [11].

Our main hypothesis in this study is that DNNs with greater capacity are more performant for SAS-ATR. To address this hypothesis, we evaluate the recognition accuracy of six different DNN architectures on a large collection of SAS-ATR data comprising three distinct geographic locations. To evaluate model accuracy, we utilize a realistic leave-one-location-out cross-validation procedure. For each DNN architecture we evaluate high-capacity and low-capacity versions, so that we can isolate the impact of capacity. Of our six DNN architectures, we use three CNN models and three transformer-based models, to isolate the impact of architecture.

While highly overparameterized DNNs often achieve greater generalization, they have also been found more challenging to optimize (i.e., train) [19] particularly in the case of transformer networks [20], [21]. A variety of methods to overcome these challenges have been developed, sometimes termed *regularization*. In pursuit of our overall goal of using high-capacity DNNs, we systematically investigate the use of several major regularization techniques for SAS-ATR: e.g., parameter pretraining, data augmentation [22], Drop Path [23] and Weight Decay [24]. Each of these parameters have hyperparameters (i.e., user-chosen parameters) that can be tailored to a specific application, and we systematically investigate good hyperparameter settings for SAS-ATR. For example, pretraining entails training the SAS-ATR model on a related dataset first where training data is much more abundant, and we investigate pretraining on several different sources: optical imagery, Synthetic Aperture Radar, and ultrasound imagery. Regarding augmentation, we investigate eleven individual augmentations and optimize their hyperparameters. We also optimize key hyperparameters associated with Drop Path and Weight Decay. We summarize our contributions as follows:

1) The first systematic evaluation of high-capacity DNNs for SAS-ATR. We evaluate six different DNN architectures, including three transformer-based models and three convolution-based models.
2) A systematic investigation of major regularization techniques for SAS-ATR. We identify appropriate adaptations of these approaches to SAS-ATR, and we find that their use substantially improves the accuracy of high-capacity DNNs for SAS-ATR.
3) We compare high-capacity and low-capacity DNN models, including some recent models from the SAS-ATR literature. We find that high-capacity models generally outperform low-capacity models, leading to state-of-the-art SAS-ATR accuracy.

The remainder of this paper is organized as follows: section II covers related work; section III discusses training DNNs for SAS-ATR; section IV presents the methods utilized in our experiments; section V and VI describe our experiments and present results; section VII presents our conclusions and future work. We also include an appendix for supplemental content.

## II. Related Work

### *A. SAS-ATR and Neural Network Capacity.*

There has been limited research into how model size relates to DNN performance on SAS datasets. One paper discussing this topic focuses on training tiny CNNs on SAS to address overfitting concerns formerly associated with large network capacity and scarce training data [10]. Recent machine learning (ML) research suggests that such concerns may be overstated [14], [15], [25], specifically due to the so-called Double Descent phenomenon. [16], This has motivated a new general approach to ML where larger networks are generally preferred for maximum test performance [17], dubbed the "overparameterized regime" [26], [27], [28]. In this work, for the first time in SAS-ATR, we adopt double descent as a design principle and focus on effectively training especially high-capacity DNN models.

### *B. CNNs and Transformers in SAS-ATR.*

CNNs have historically been the method of choice for SAS-ATR [11]. Having been found superior to the formerly used feature-based classifiers [6] there has been much research into the use of CNNs for this application. Early adoption of CNNs for SAS-ATR included their use for semantic segmentation of the seafloor in a semi-supervised manner [29], as ATR networks that ingest multiple representations of SAS data [30], and as phase-informed classifiers that leverage the complex nature of SAS data to bolster performance [31], among others. Since these early innovations of CNN-based SAS-ATR there has been further work in improving these systems, such as the incorporation of wavenumber domain partitioning in training data [7], the inclusion of structural priors to diminish false alarm rates [32], and insights into the effects distribution shifts of SAS data have on CNN performance [33].

More recently Vision Transformers (ViT) [34] and its extensions [35], [36] have led to state-of-the-art performance on general CV tasks, and some studies have explored them for SAS-ATR. CNN/transformer hybrid networks have been used for volumetric target classification in 3D SAS data [37], and detection methods leveraging the attention mechanism from transformer-based networks have been investigated for SAS data [38]. A comparison for ATR in side-scan sonar (SSS) seems to suggest transformer-based networks can outperform CNNs on sonar ATR tasks [39], [40]. While transformers are now being investigated, they remain underutilized in SAS-ATR [11]. In this work we make the first systematic comparison of CNN and transformer-based models for SAS-ATR.

### *C. Pretraining Neural Networks for SAS.*

While there is research that suggests transfer learning between SAS-ATR networks with different sensors is superior to "from scratch" training [41], the same work disputes whether transferring from networks trained on optical imagery has the same benefits. On the contrary, other research has shown that transferring weights from networks trained on optical imagery can improve SAS-ATR performance [42]. Beyond strictly optical imagery, there has been work that suggests pretraining on other imaging modalities such as ultrasound and synthetic aperture radar (SAR) data offers some performance improvements when applied to SSS classification [43], although no such study exists for SAS-ATR. Further comparisons of pretrained model performance in SAS applications have seen very limited investigation. Considering the lack of consensus on the utility of models pretrained on non-sonar imagery for SAS-ATR, this question requires further investigation. In this work we compare the performance of CNN and transformer-based networks using pretrained weights from SAR, ultrasound imagery, and optical imagery against a randomly initialized baseline.

### *D. SAS Data Augmentation.*

Given the lack of labeled SAS training samples there has been some research on applying data augmentation to SAS-ATR. There are two categories of augmentation when it comes to SAS data: those that preserve SAS's unique scene geometry and physical properties (i.e. "physical" augmentations) and those that do not ("non-physical" augmentations). Physical augmentations have seen more investigation with authors finding the use of small affine transformations [6], [10], and the incorporation of spectral information [7], [31] to offer performance improvements in SAS-ATR. Non-physical augmentations, however, have also been found to hold potential benefits in more recent years with two separate studies finding non-physical augmentations to elicit even higher performance gains than their physical counterparts [44], [45]. Given these findings it seems well established that data augmentation is helpful for improving SAS-ATR performance. In this work we perform the first large-scale and systematic evaluation of augmentation on networks of varying capacities and architectures.

### *E. Network Regularization in SAS-ATR.*

Research into the regularization of SAS-ATR networks is not new. Perhaps the most direct contributions to this inquiry has been work on the incorporation of SAS-specific structural priors as regularization terms in loss functions [32] as well as statistical regularizers that urge favorable statistical patterns onto weight values [46]. The former's direct incorporation of SAS domain knowledge into a CNN's learning process was found to improve performance beyond that of other performant SAS-ATR networks of the time, and the latter's enforcement of statistical independence between weight values was found to benefit performance. Neither of these works, however, includes a controlled comparison of how such regularization compares to traditional CV methods of weight decay [24], DropPath [23], and data augmentation [22], nor do they compare how networks of varying architectures and capacities are impacted by these methods. In this work we provide the first systematic comparison of the impacts of weight decay, DropPath, and data augmentation on six network architectures (three CNN and three transformer-based) with multiple capacity variants.

## III. Problem Setting

The goal of our SAS-ATR problem is to classify real-valued beamformed patches of SAS-ATR data into one of two labels: target or non-target. More formally, we wish to assign a binary label $y \in \{0,1\}$ to some beamformed SAS data, $x \in \mathbb{R}^{h\times w}$, indicating whether it corresponds to some pre-defined object of interest ($y = 1$), or *target,* or not ($y = 0$). In our context the data $x$ corresponds to the magnitude component of a beamformed SAS image. We assume that data is drawn from some unknown distribution $(x_i, y_i) \sim P_{XY}$ and goal is to find some mapping of the form $y = f(x)$ that achieves low expected classification error, given by **Eq. (1)**.

$$f^* = argmin_f\ \mathbb{E}_{\sim P_{XY}}[\mathcal{L}(f(x), y)] \qquad (1)$$

where $\mathcal{L}$ is some measure of error, or *loss*, between the prediction $\hat{y} = f_\theta(x)$ and the ground truth $y$ (e.g., cross-entropy). In practice we approximate **Eq. (1)** empirically by minimizing an *empirical loss*, given by **Eq. (2)**

$$f^* = argmin_f \sum_{(x_i, y_i) \in D} [\mathcal{L}(f(x_i), y_i)] \qquad (2)$$

where $D = (x_i, y_i)_{i=1}^{I}$ is a dataset of i.i.d. samples from $P_{XY}$. This problem setup is typical within SAS-ATR [5], [6], [7], [9], [10], [11] etc. Our work investigates the use of data-driven models to solve this problem, in which we employ a parameterized function $y = f_\theta(x)$, where $\theta$ are parameters that influence the mapping between $x$ and $y$. Then our data-driven modeling problem takes the form of **Eq. (3)**

$$\theta^* = argmin_\theta \sum_{(x_i, y_i) \in D} [\mathcal{L}(f_\theta(x_i), y_i)] \qquad (3)$$

where our goal is now to find the parameters that minimize our empirical loss instead of the function.

## IV. Methods

This section provides a brief introduction to the methods utilized in our experiments, and their most essential details, such as key hyperparameters (i.e., parameters set by the designer) that we manipulate in our experiments.

### A. Deep Neural Networks (DNNs)

In this work we utilize DNNs to solve the SAS-ATR problem posed in Sec. III. DNNs are a broad class of parameterized functions of the form $y = f_\theta(x)$, where the parameters $\theta \in \mathbb{R}^p$ influence the input-to-output mapping. It has been shown that, with the proper settings of $\theta$, DNNs can accurately classify SAS-ATR patches. The process of identifying good parameters $\theta$ is called *training* and is discussed in Sec. IV-B.

The effectiveness of a DNN for a given problem, such as SAS-ATR can vary substantially based upon its design. We discuss here two key design choices that we will utilize and vary in our experiments: model *architecture*, and model *capacity*. The DNN *architecture* refers to the functional form of $f_\theta$, which can strongly influence its efficacy for a particular problem [34], [47]. Most modern DNNs consist of multiple functional *layers* of the form $f_\theta = f_{\theta_L} \circ \ldots \circ f_{\theta_2} \circ f_{\theta_1}(x)$, where $\theta_l \in \mathbb{R}^{p_l}$ represents the parameters associated with layer $l$. The number of layers $L$ is typically called the network *depth,* and $p_l$ typically called the *width* of the $l^{th}$ network layer. DNN *capacity* is defined by the total number of parameters, $p$, which can be mediated by varying the depth and width of a network. The individual layers in a DNN often have different functional forms, often suited to a particular purpose, or type of data. Two widely utilized DNN layers are convolutional layers, and transformer layers. Note that these layer types correspond to general classes of layer structures that have shared properties, rather than just one specific structure.

A DNN *architecture* is defined by a specific combination of layer types: often a combination that has been found effective for certain problems, such as image recognition. Modern DNN architectures are often dichotomized into *Convolutional Neural Networks* (*CNNs*), which primarily comprise convolutional layers, and *Transformer Networks (Transformers),* that primarily comprise transformer layers. Both CNNs and Transformers have been found highly effective for processing imagery data, depending upon the context. A full discussion of these architectures is beyond the scope of this paper, however, we highlight two key differences. In principle, transformers more easily learn relationships between distant input values (e.g., on opposite sides of the input, $x$) than convolution. Second, transformers employ "attention" processing, whereby the importance assigned to each portion of the input can easily vary, relative to CNNs [34].

**Architecture and Capacity Choices.** In this work we compare the SAS-ATR performance of six different DNNs: three CNNs (ResNet [47], ResNeXt [48], and ConvNext [49]) and three transformers (ViT [34], SWIN [35], and HiViT [50]). To evaluate the impact of network capacity we vary the capacity for each architecture, resulting in 14 total networks, as summarized in **Table 1**.

### B. DNN Training

Here we describe the process of finding good parameters, $\theta$, which is fundamentally achieved by solving **Eq. (3)** with a DNN model. A solution to **Eq. (3)** is typically found using some type of gradient descent algorithm, which uses the gradient of the loss with respect $\boldsymbol{\theta}$ to iteratively adjust $\boldsymbol{\theta}$ in a way that reduces the empirical loss over the data $\boldsymbol{D}$. A single iteration of gradient descent training takes the generic form

$$\boldsymbol{\theta} \leftarrow \boldsymbol{\theta} - \lambda \nabla_{\boldsymbol{\theta}} \mathcal{L}(f_{\boldsymbol{\theta}}(x_i), y_i) \qquad \textbf{(4)}$$

where $\nabla_{\boldsymbol{\theta}}$ denotes the gradient operator, and $\lambda$ is a hyperparameter (i.e., set by the designer) that controls the relative magnitude of change made to $\boldsymbol{\theta}$ in each iteration.

**Training Optimizer Choice.** There are several widely-used variations of gradient-based training, such as Stochastic Gradient Descent [51], Adam [52], and AdamW [53] optimizers. In this work we will use the AdamW optimizer, which introduces two key hyperparameters: minibatch size $\beta$, and weight decay rate, $\tau$. **Eq. (4)** only ingests a single data sample, but in AdamW the gradients are computed over $\beta$ training data samples, and then averaged, before updating $\boldsymbol{\theta}$**.** It is termed an "epoch" once every data sample in $D$ has been utilized in a minibatch, and training often continues for several epochs. The weight decay in AdamW is a form of regularization and will be discussed further in Sec. IV-C.

**Loss Function Choice.** One other training consideration is the choice of loss function $\mathcal{L}$. In classification settings such as ours, it is most common to use the cross-entropy loss. However, we have far more non-target training instances, many of which are easily classified. To address this, we utilize *focal loss* [54] (**Eq. (7)**), which modifies the cross-entropy loss to reduce the influence of abundant easily-classified input, as we have. Focal loss uses two hyperparameters: the focusing parameter (γ) that reduces loss on correctly classified samples and the class balance parameter (α) that magnifies the loss for predictions made on samples from the minority class. For our experiments we use γ = 2.0 and α = 0.25. Further information about focal loss can be found in subsection[1]D of the appendix.

**Performance Evaluation**. Once training is complete, it is useful to evaluate the classification accuracy of the trained DNN and compare it to other competing trained models. It is well-established that the data used to train the model – the training data- often yields optimistic accuracy estimates, making it unsuitable for evaluating models. Therefore, it is common to withhold two subsets of the data, $D$, from training so that they can later be used for model evaluations. One set, termed the "validation" set is often used to monitor the progress of training. Another set, the "testing set", is utilized to estimate real-world performance, or compare performance of competing models. In this work we utilize a systematic procedure for splitting up the available data, termed *k-fold cross-validation* [55], which is designed to make efficient use of available data when training and evaluating models. This process splits the data into $k$ disjoint sets, each of which serve once as a testing set, while the remaining folds are aggregated into a training (and/or validation) set.

**Performance Metrics**. In this study we report our results using two metrics: Average Precision (AP) [56] and Optimal F1. F1-score is defined as the harmonic mean of precision and recall. To obtain a precision and recall, it is necessary to apply a decision threshold, $t$, to the classifier output, such that all predictions greater than $t$ are assigned a label $y = 1$, and $y = 0$ otherwise. Typically, $t = 0.5$, however, this can result in pessimistic F1 scores for poorly calibrated classifiers, such as those using Focal Loss, as in our experiments. Optimal F1 refers to the F1 score obtained when we select the setting of $t$ that maximizes F1. This results in a score that is more resilient to poorly calibrated classifiers.

### *C. DNN Regularization*

Although high-capacity DNNs often perform better, they can still suffer from high variance between training data points and thereby fail to generalize well to test data. It has been found helpful to use regularization techniques to mitigate this tendency, which encourages the model to interpolate more effectively (e.g., smoothly) between training data points. In this work we investigate several regularization techniques: pretraining, DropPath, weight decay and data augmentation. We discuss each approach, and key hyperparameters that we optimize in our experiments.

**DropPath [23]** involves muting network connections by sampling from a Bernoulli distribution using a hyperparameter $p \in (0,1]$ as outlined in **Eq. (5)**.

$$\boldsymbol{r_i \sim \mathbf{Bernoulli(p)}} \qquad (5)$$

When $r_i$ is zero this path is dropped, preventing its output from contributing to the final representation. This encourages generalizability by preventing parameters from adjusting too heavily to fit training data. Our experiments in Sec. VI-C vary the hyperparameter p to modulate the aggressiveness of Drop Path and evaluate how these changes impact performance. DropPath (also known as "stochastic depth") is based on traditional dropout [57], however DropPath has become the more favored due to superior empirical performance.

**Weight Decay[24].** Weight decay forces the parameters of the networks to shrink towards zero in each iteration of training, which has been found to result in smoother interpolation between training points. To implement weight decay we use the Adam-W optimizer [53], which modifies the widely-used Adam optimizer [52] to include weight decay.

**Pretraining** involves initializing networks with weights from networks that have already been trained on other data. Compared to training "from scratch" (i.e. not using pretrained weights) this is typically preferred as it allows networks to be initialized with some understanding of spatial phenomena (edges, corners, blobs etc.) which can be fine-tuned to fit the training data. In CV, the use of pretrained weights from ImageNet [58] is generally regarded as widely beneficial to performance, however in SAS-ATR there are lingering questions as to whether such weights from networks trained on optical imagery provide similar such benefits, or if it would be more advantageous to use weights from networks trained on an imaging modality closer to SAS (such as SAR). We address these questions in this work.

**Data Augmentation [22].** Data augmentation is the practice of creating novel representations of data samples by transforming them. This typically improves a network's generalizability by providing unique samples from existing ones, effectively simulating unseen samples by making changes to the data. For a data sample $x_i$ a given augmentation $a(x_i)$ is applied based on a probability parameter $p_a \in [0,1]$ replacing $x_i$ with the sample $\hat{x}_i$.

$$\boldsymbol{\hat{x}_i \leftarrow m_i a(x_i) + (1 - m_i) x_i} \\ \boldsymbol{where\ m_i \sim \mathbf{Bernoulli}(p_a)} \qquad (6)$$

Historically in SAS-ATR literature there is some discussion about the superiority of "physical" over "non-physical" augmentations with the former being augmentations that produce images that preserve the essential physical properties of SAS images (i.e. platform track orientation, acoustic shadow, scattering characteristics etc.) and the latter being those that do not. In more recent years results have shown that both can be beneficial to performance [44], [45].

## V. Main Experiments

We conduct experiments that seek to address two main questions for SAS-ATR: (i) do high-capacity DNN models tend to outperform smaller models; and (ii) do transformer-based architectures outperform convolutional ones?

### *A. Experimental Design Details*

To address question (i) above, we evaluate SAS-ATR performance of multiple DNN architectures, respectively, as we vary their capacity. SAS-ATR performance is evaluated in terms of binary classification accuracy, where the model must classify beamformed SAS images into those representing a target, or not. We train and evaluate each model using the same procedure, which includes optimization of several hyperparameters to help ensure that each model achieves its full potential performance. This design is intended to isolate the impact of capacity from that of model architecture. To address question (ii) we conduct these experiments with several different transformer and convolution-based architectures.

**Experimental SAS Dataset.** Our dataset consists of 500,000 beamformed SAS image patches, drawn from three different geographic locations with unique oceanic and bathymetric conditions. The patches comprise regions of interest selected from a larger beamformed images of the seafloor using an energy-based detector. This process is depicted in **Fig 3**. Each extracted patch is the same size, but they are later resized to a height and width of 256 pixels for ingestion by deep learning models. Each patch is independently mapped to 8-bit intensity values by mapping the minimum intensity in each patch to 0 and the maximum to 255. Each geographic region comprises a similar number of targets, but very different numbers of non-targets. In preparation for experiments, we randomly subsample, or replicate, the non-target patches so that each location contains ~18500 images and thereby approximately a 30:7 clutter to target ratio. This ratio was chosen to achieve greater balance between target and non-target data, but without omitting too much non-target data.

**Model Training and Performance Assessment**. To provide a real-world performance assessment, we use a challenging leave-one-region-out cross-validation scheme. Each model was trained for 40 epochs using the ADAM-W optimizer [53] and a learning rate schedule consisting of a linear warmup from 1e-5 to 1e-4 for the first ten epochs and cosine annealing for the remaining epochs. To help ensure this training schedule was sufficiently-long for all models to converge, we monitored training loss of all models in **Table 1** and found every model reached near-perfect classification of the training set (i.e., a loss of $\leq 1e-3$). In similar fashion we used a validation set to verify that none of the networks were overfitting. The validation set consists of ~160,00 SAS images with a 1:159 ratio of targets to non-targets. Performance of each model was evaluated on the test folds using Average Precision (AP) and Optimal F1 score (Opt. F1), as discussed in Sec. IV-B. **Fig 4** depicts a high-level overview of this process.

**Hyperparameter and Regularization Optimization**. We perform a thorough hyperparameter optimization for each model (i.e., each combination of an architecture and its capacity) to help ensure that it reaches its full potential performance. As discussed in Sec. IV-C, effective regularization strategies are often important to realize the benefits of high-capacity models studied here. We sequentially considered and optimized several forms of regularization for each model in our study: weight pretraining, Drop Path, weight decay, and data augmentation (in this order). In each step of the optimization, we choose the settings that leads to the best performance, as evaluated by the average optimal F1-score across each of the three folds in our cross-validation scheme. We first considered use of ImageNet pretrained weights, versus training from scratch. We then jointly search over combinations of weight decay of 5%, 10% and 20% and the Drop Path rates of 10%, 20% and 30%. Lastly, we evaluate whether each model benefited from inclusion of a SAS-optimized data augmentation. Sec. VI-B describes the augmentation policy and the process used to optimize it for SAS.

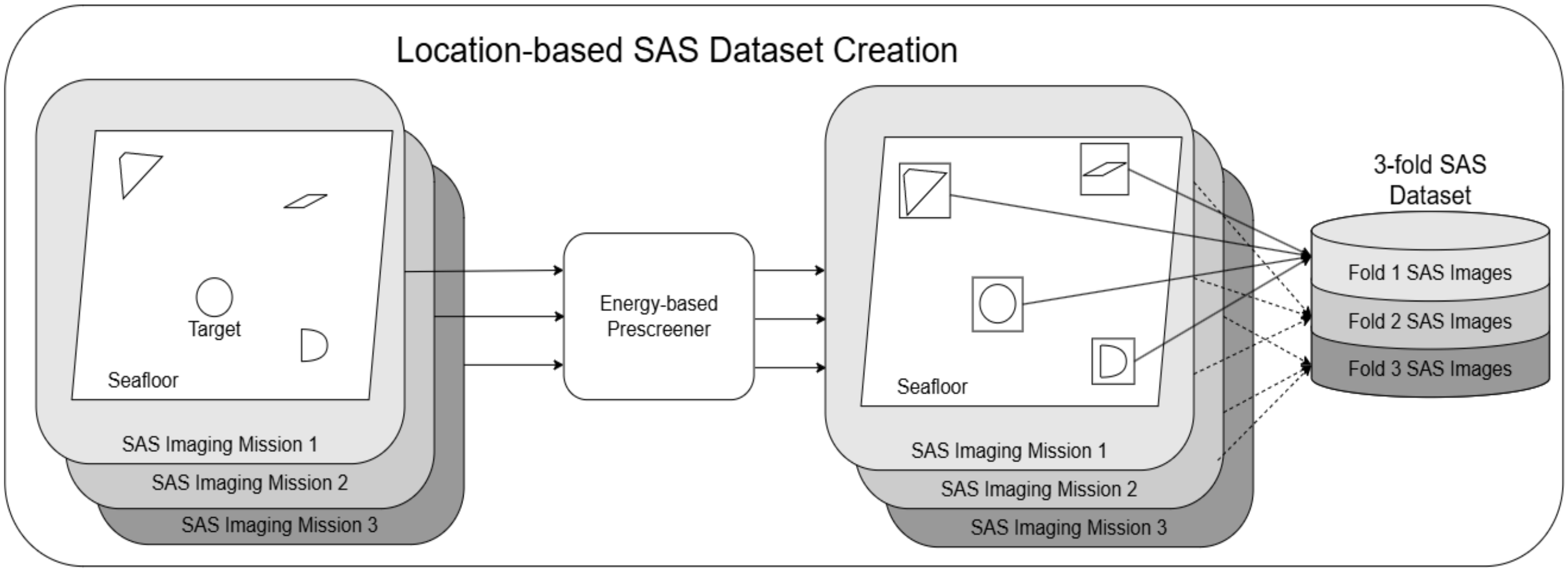

**Fig 3: Process used to create our location-based 3-fold SAS Dataset. Three missions covering different geographical regions of seafloor are conducted to create SAS images which are fed to an energy-based prescreener to produce samples containing objects of high reflectivity.**

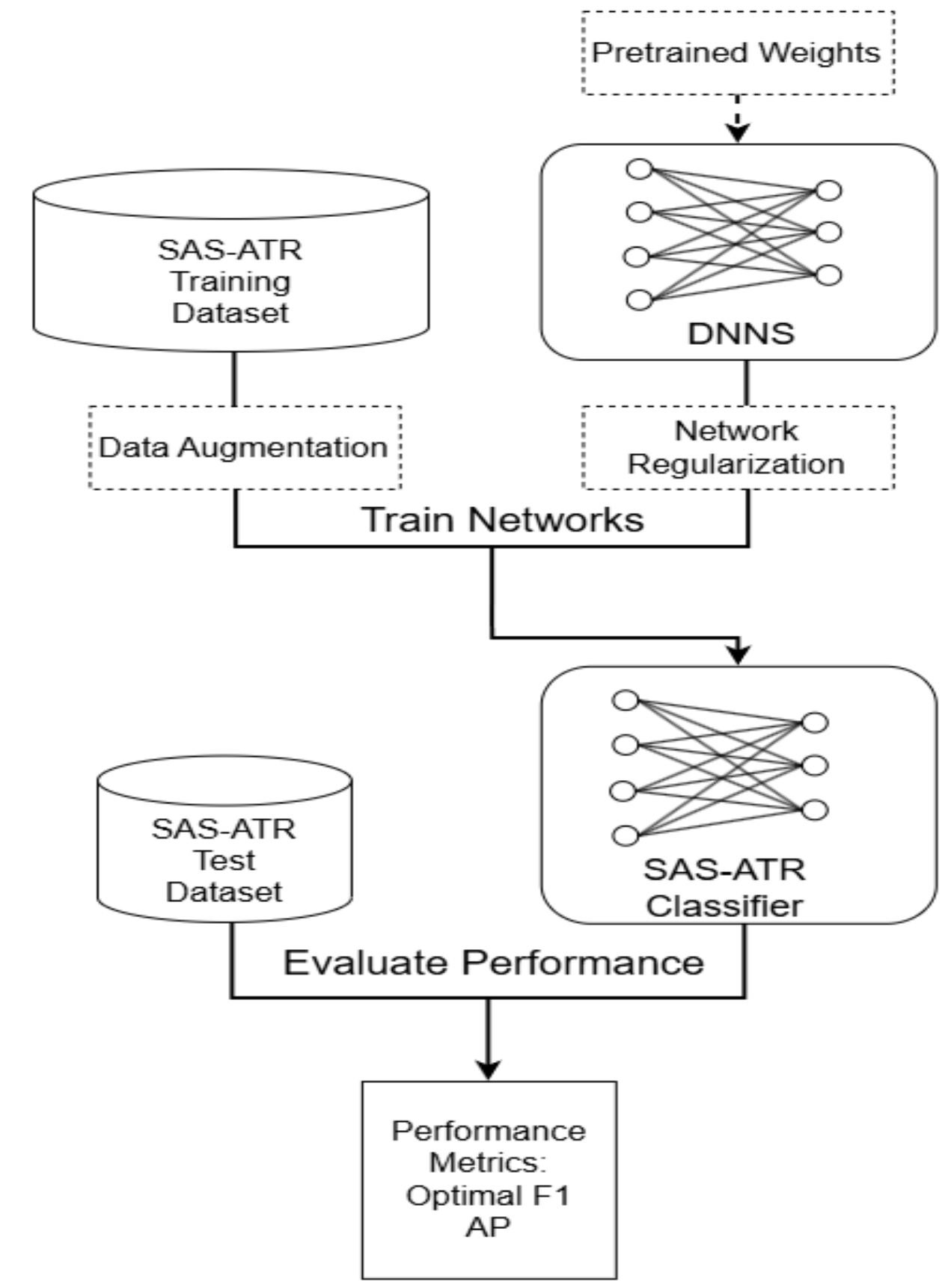

**Fig 4: SAS-ATR classifier training process. The use of data augmentation, network regularization, and pretrained weights (all in dotted boxes) are each methods of improving classifier performance that we experiment with in Sec. VI.**

### *B. Experimental Results*

Our main results are reported in **Table 1** and we next discuss each of our major findings with respect to the results.

**Impact of model capacity.** The results indicate that, given a fixed architecture, model performance generally increases (often substantially) as model capacity increases. For example, the ResNeXt-152 improves by 10% AP over the ResNeXt-50, leading to the highest AP among convolutional models. The Swin-Base improves 10% AP over the Swin-Tiny model, leading to the highest performance among transformer models. These results suggest the benefits of capacity are not unique to convolution or transformers. However, the performance impact does vary in general, with some models exhibiting little or no improvement. For example, the HiViT and ResNet models exhibit more modest improvements, or even a (slight) performance degradation in one case. Importantly, while capacity increases do sometimes lead to degradation, they are usually very small compared to the improvements often yielded, and so there is little risk of reducing performance by using higher-capacity models.

**Comparison with Existing SAS-ATR Models.** Existing SAS-ATR classifiers are low-capacity models. Here we train two models adopted from the literature and compare them with the models in our experiments: a ResNet utilized in two recent studies [13] [32], and the "TinyCNN" [10]. To ensure a fair comparison between networks we use the same training/optimization scheme used for our other models, as outlined in Sec. Main Experiments-A. Our results indicate that every one of our proposed high-capacity models outperforms these two existing SAS-ATR networks, with many of our proposed models having substantially greater performance.

| Network | | Variant | Capacity ($\times 10^6$) | AP | Opt. F1 |
|---|---|---|---|---|---|
| Convolutional | ResNet | 50 | 24 | 79.52 | 88.38 |
| | | 101 | 43 | 82.71 | 90.05 |
| | | 152 | 58 | 82.54 | 90.29 |
| | ResNeXt | 50 | 23 | 79.94 | 88.93 |
| | | 152 | 58 | 89.54 | 90.04 |
| | ConvNext | Tiny | 28 | 57.19 | 74.60 |
| | | Base | 88 | 84.36 | 91.17 |
| Transformers | ViT | Small | 22 | 67.62 | 81.80 |
| | | Base | 86 | 67.95 | 81.60 |
| | | Large | 303 | 67.07 | 80.80 |
| | SWIN | Tiny | 28 | 73.30 | 85.93 |
| | | Base | 87 | 84.69 | 92.24 |
| | HiViT | Small | 37 | 67.94 | 84.07 |
| | | Base | 78 | 69.84 | 84.41 |
| TinyCNN [10] | | | <1 | 55.41 | 71.02 |
| ResNet18 [13] [32] | | | 11 | 56.96 | 69.87 |

**Table 1. Average Precision and Optimal F1 Scores for each model, categorized as a Transformer or a Convolutional model type. Models that we implemented from the literature include a reference. The Capacity of each model is rounded. Bolded performance is the best within its category: transformer or convolutional.**

**Best Overall Performing Model?** While our results do not suggest a singular best performing SAS-ATR model there are network architectures that show superior performance. SWIN, ResNeXt and ConvNeXt are the most performant networks tested in this work, with their largest variants having the best performance. This suggests the superiority of these methods over alternatives for SAS-ATR, which is consistent with broader computer vision literature where SWIN and ConvNeXt often obtain state-of-the-art performance on natural imagery.

## VI. Additional Analysis

In this section we report additional experiments to isolate the impact of different modeling choices.

### *A. Impact of Pretraining*

Pretraining has been found to be most helpful when it is done with data from a similar modality. Therefore, we perform a controlled comparison between pretrained parameters from three imaging modalities, including some that are potentially more qualitatively similar to SAS. To determine the optimal modality upon which to pretrain SAS-ATR networks we compared pretraining on ImageNet (optical), SAR (radar), and Ultrasound imagery (acoustic). In these experiments we only use ResNet and ViT networks because we could not find all of the necessary pretrained weights for other networks. Beyond this, all other training procedures used in these experiments are identical to the experiments in Sec. Main Experiments. Our results are reported in **Table 2** and indicate that pretraining is generally beneficial compared to training from scratch. Furthermore, ImageNet pretraining was always superior compared to other pretraining strategies. Based upon these results, we conducted a second experiment wherein we trained a large number of our models from scratch, as well as with ImageNet pretraining. The results of this experiment are reported in **Table 4** and show that ImageNet pretraining is beneficial (often substantially) for every single model, and by both of our performance metrics. These results suggest that ImageNet pretraining is consistently beneficial for SAS-ATR, compared to training from scratch, or compared to other pretraining.

The superiority of ImageNet weights compared to others may be surprising, given the greater apparent visual similarity of SAR or Ultrasound data to SAS. We hypothesize that the superiority of ImageNet weights may be explained by the substantially greater size and diversity of the ImageNet dataset, and greater relative attention given to ImageNet models within the vision community. It is plausible that pretraining on other modalities (e.g., SAR) would be superior for SAS-ATR if the advantages of ImageNet weights were mitigated, however, this is challenging, and beyond the scope of this work.

*B. Impact of Augmentation*

Data augmentation has recently been found beneficial for training DNN-based SAS-ATR models [44], [45]. We seek to leverage these recent findings to help regularize our high-capacity models, and to generally improve performance of the models in our study. To do this we consider ten different individual augmentations recently found to be beneficial for SAS [45], and search for a combination of them, termed a *policy*, that performs well for our SAS ATR data. Each individual augmentation possesses one (or more) hyperparameters, which can strongly impact its performance on SAS-ATR [45]. For each augmentation we measure its performance as we vary its hyperparameter settings to identify the best one. We perform this search separately for one convolutional model (ConvNeXt) and one transformer model (SWIN), since it has been found that the best hyperparameters may depend upon the DNN architecture [45]. The results of this search indicate that ConvNeXt was only able to achieve marginal performance improvements with optimized hyperparameters while SWIN saw consistent improvements. The hyperparameter values found to be optimal are shown in **Table 7** of the appendix.

Following hyperparameter optimization, we conduct a greedy sequential search (SS) to construct an augmentation policy that maximizes performance. Performance comparisons of SWIN-base and ConvNeXt-base with no augmentations, the best single augmentation, and SS policy are reported in **Table 4** and

| Network | Pretraining | AP | Opt. F1 |
|---|---|---|---|
| ResNet50 | None | 57.38 | 71.98 |
| | ImageNet | 67.84 | 78.46 |
| | SAR | 65.74 | 76.75 |
| | Ultrasound | 62.79 | 75.07 |
| ResNet101 | None | 54.40 | 68.81 |
| | ImageNet | 73.69 | 84.09 |
| | SAR | 65.53 | 76.62 |
| | Ultrasound | N/A | N/A |
| ViT-b | None | 57.98 | 73.11 |
| | ImageNet | 64.27 | 76.73 |
| | SAR | 57.37 | 71.10 |
| | Ultrasound | 57.75 | 71.34 |
| ViT-l | None | 53.98 | 71.54 |
| | ImageNet | 63.00 | 73.98 |
| | SAR | 55.55 | 68.69 |
| | Ultrasound | 51.88 | 72.85 |

**Table 2: Results of each ViT and ResNet variant using pretrained weights from networks trained on each different imaging modality.**

| Network | | Variant | Capacity ($\times 10^6$) | AP | Pretrained AP | Opt. F1 | Pretrained Opt. F1 |
|---|---|---|---|---|---|---|---|
| Convolutional | ResNet | 50 | 24 | 57.38 | 67.84 | 71.98 | 78.46 |
| | | 101 | 43 | 54.40 | 73.69 | 68.81 | 84.09 |
| | | 152 | 58 | 54.43 | 74.40 | 70.67 | 85.20 |
| | ResNeXt | 50 | 23 | 57.59* | 74.83 | 71.77* | 84.27 |
| | | 152 | 58 | 53.16 | 71.42 | 71.06 | 82.06 |
| | ConvNext | Tiny | 28 | 19.36 | 49.73 | 32.06 | 63.96 |
| | | Base | 88 | 19.7 | 78.63 | 32.92 | 87.69 |
| Transformers | ViT | Small | 22 | 50.69 | 52.03 | 68.45 | 64.50 |
| | | Base | 86 | 57.98* | 64.27 | 73.11* | 76.73 |
| | | Large | 303 | 53.98 | 63.00 | 71.54 | 73.98 |
| | SWIN | Tiny | 28 | 57.10 | 57.61 | 70.82 | 71.13 |
| | | Base | 87 | 52.22 | 75.88 | 68.14 | 87.31 |
| | HiViT | Small | 37 | 53.56 | 53.70 | 63.43 | 65.09 |
| | | Base | 78 | 54.66 | 57.51 | 65.64 | 66.84 |

**Table 3. Average Precision and Optimal F1 Scores for each network and their variants. Results are divided by CNNs (top), transformer-based networks (bottom), non-pretrained networks (left) and pretrained networks (right)**

| SWIN-base | AP | Opt. F1 |
|---|---|---|
| No Augmentations | 75.88 | 87.31 |
| Best Single Aug. (Zoom) | 83.18 | 90.62 |
| End of Sequential Search | 85.05 | 92.28 |

**Table 4. Results for greedy sequential search of augmentation policy for SWIN-base**

| ConvNeXt-base | AP | Opt. F1 |
|---|---|---|
| No Augmentations | 78.63 | 87.69 |
| Best Single Aug. (Zoom) | 79.05 | 86.90 |
| End of Sequential Search | 79.05 | 86.90 |

**Table 5. Results for greedy sequential search of augmentation policy for ConvNeXt-base**

**Table 5**. Our results indicate that ConvNeXt-base receives only a marginal improvement in AP with the inclusion of Zoom. Additional augmentations provided no improvement. This is contrasted with SWIN-base which benefitted from almost every augmentation and was able to achieve a near 10-point gain in AP and 5 points in optimal F1-score by including both Zoom and Rotation augmentations. These results provide some evidence that transformers may benefit more from augmentation and ultimately provide better performance with more training data and/or regularization. This is consistent with the broader computer vision literature [59], [60].

### *C. Impact of Regularization*

Here we evaluate the aggregate impact of regularization on our models. In these experiments, "regularization" comprises weight decay, DropPath, and data augmentation. The regularization hyperparameters were optimized in the manner described in Sec. Main ExperimentsA. We evaluate the performance of models with regularization, and without, respectively. All models in these experiments include ImageNet pretraining. **Table 6** compares the performance of networks with and without regularization. **Table 9** in the appendix details which hyperparameter values were found to elicit the strongest performance for each network. Regularization was found to be generally helpful as it improved the performance of each network. The magnitude of such improvements was consistent across architectures; however smaller networks often saw the most substantial improvements. These networks were still outperformed by their larger counterparts.

## VII. Conclusions

In this work we examine the performance impacts of network capacity, network architecture, and regularization on DNN models for SAS-ATR. To understand how architecture impacts performance we examined six popular classification architectures (three transformer-based and three CNNs) under controlled optimization and evaluation regimes and compared their performance in terms of average precision and optimal F1-score. We also included two recent DNN models from the SAS-ATR literature in our comparisons. Our main findings are as follows:

- Increasing network capacity usually improves SAS-ATR performance, often substantially. It is never significantly detrimental.
- Model architecture has a significant impact on SAS-ATR performance. The best overall performance was achieved by SWIN, ResNeXt and ConvNeXt models with high-capacity variants. These models substantially outperformed the two baseline models from the SAS-ATR literature.
- Regularization is crucial for realizing the advantages of increasing model capacity.
- Regularization using pretrained ImageNet parameters was always beneficial compared to random model parameter initialization. Pretraining on ImageNet was always superior to pretraining on SAR or Ultrasound data.

| Network | Variant | AP – No Reg. | AP – Best Reg | Opt. F1 – No Reg. | Opt. F1 – Best Reg. |
|---|---|---|---|---|---|
| ResNet | 50 | 67.84 | 79.52 | 78.46 | 88.38 |
| | 101 | 73.69 | 82.71 | 84.09 | 90.05 |
| | 152 | 74.40 | 82.54 | 85.20 | 90.29 |
| ResNeXt | 50 | 74.83 | 79.94 | 84.27 | 88.93 |
| | 152 | 71.42 | 89.54 | 82.06 | 90.04 |
| ConvNext | Tiny | 49.73 | 57.19 | 63.96 | 74.60 |
| | Base | 78.63 | 84.36 | 87.69 | 91.17 |
| ViT | Small | 52.03 | 67.62 | 64.50 | 81.80 |
| | Base | 64.27 | 67.95 | 76.73 | 81.60 |
| | Large | 63.00 | 67.07 | 73.98 | 80.80 |
| SWIN | Tiny | 57.61 | 73.30 | 71.13 | 85.93 |
| | Base | 75.88 | 84.69 | 87.31 | 92.24 |
| HiViT | Small | 53.70 | 67.94 | 65.09 | 84.07 |
| | Base | 57.51 | 69.84 | 66.84 | 84.41 |

**Table 6. Best Average Precision and Optimal F1 Scores for each network using no regularization and the best regularization scheme found in this study. Each network is pretrained with ImageNet weights.**

- Regularization with DropPath and weight decay was generally beneficial
- Regularization with data augmentation was highly beneficial for transformers, but less-so for convolutional models.

Collectively our results suggest that high-capacity DNNs, when properly regularized, represent a promising methodology for SAS-ATR. Our results here provide insight into how to train and properly regularize such models to maximize their performance on SAS-ATR.

**Limitations and Future Work.** We have limitations regarding our pretraining conclusions as we only consider transfer-style pretraining. While ImageNet pretraining was found universally helpful, we cannot determine if it is truly the optimal pretraining method as we do not compare other methods such as self-supervised learning. Future work should be done to compare ImageNet pretraining to self-supervised learning to determine which is truly best for SAS-ATR. There are further opportunities for future work that we have not yet outlined. It is unclear from our experiments exactly how network architecture type (transformer-based or CNN) interacts with other aspects of training in SAS-ATR. We find no significant trends to such an effect, suggesting that more work should be done to compare CNNs and transformer-based networks for SAS-ATR as our results imply that both are viable, but neither is definitively superior.

# VIII. APPENDIX

## *A. Non-SAS Pretrained Weights*

All ImageNet weights are sourced from Torchvision [61]. Hugging Face [62] is used to source ultrasound weights and the SAR weights we use for our ResNet models are from an open-source project with the goal of providing such weights for SAR applications [63] which they make publicly accessible on GitHub. For the sake of replicability, **Table 8** includes links to these weights.

## *B. Training Software and Acceleration Hardware*

All training in this study was conducted using the MMPretrain package [64]. We perform data parallel training on a cluster of 4 NVIDIA A100X GPUs totaling 320GBs of high bandwidth memory for accelerated training.

## *C. Augmentation Techniques*

For replicability we include the 10 augmentations we use as well as the hyperparameters that we jointly optimize using a grid-search in **Table 7**. Each augmentation comes from the Torchvision or MMPretrain library except for the speckle noise augmentation which is implemented using details from [65]. **Table 7** also contains the optimized hyperparameter values that were found using SWIN-base. We use this network as it saw the highest performance using augmentation out of all networks tested.

## *D. Focal Loss*

Focal Loss improves this by introducing a parameter α to address class imbalance and γ to minimize the impact of trivial predictions.

$$\boldsymbol{FL = -\alpha(1 - p_{correct})^{\gamma} log(p_{correct})} \quad (7)$$

By using a weight of α when making predictions on samples from the negative class and (1-α) for samples from the positive class, an α value of 0.25 (what we use) forces loss values of predictions on targets to be more impactful to learning than those of non-targets. The use of γ allows for high-confidence predictions to have minimal impact on training. When the network makes high-confidence predictions that are correct (e.g. $p_{correct} = 0.99$) a γ greater than 1 will cause the resulting loss to be much lower than it otherwise would be, thereby minimizing the impact of such predictions on training.

| Augmentation | Key Hyperparameter(s) | Optimal Hyperparameter Value(s) |
|---|---|---|
| Contrast Shift | contrast_factor - contrast scaling factor | contrast_factor = 1.0 |
| Color Jitter | contrast – range of scaling factors for random contrast shifts<br>brightness – range of scaling factors for random brightness shifts | contrast = 0.25, brightness = 0.25 |
| Speckle Noising | rate – controls decay rate of exponential distribution from which noise is sampled<br>filterSize – defines size of square median filter | filterSize = 3, rate = 4 |
| Gaussian Blur | kernel_size – size of blur kernel<br>sigma – standard deviation of kernel | kernel_size = 3, sigma = 1 |
| Zoom | scale – range of potential scaling | scale = [0.5, 1.35] |
| Center Shift | translate – range of translation in x and y axes | translate = [-1.0, 1.0] |
| Rotation | degrees – range of potential clockwise and counterclockwise rotations. | degrees = [-40, 40] |
| Mixup | alpha – controls mixing ratio | alpha = 0.2 |
| ResizeMix | alpha – controls mixing ratio | alpha = 0.5 |
| CutMix | alpha – controls mixing ratio | alpha = 0.2 |

**Table 7. Each augmentation tested in this work, their respective hyperparameters, and the values found to elicit the strongest performance from our most performant network (SWIN-base).**

| Network | ImageNet | SAR | Ultrasound |
|---|---|---|---|
| ResNet50 | MMPretrain https://mmpretrain.readthedocs.io/en/latest/papers/resnet.html | XAI4SAR https://github.com/XAI4SAR/SAR-HUB/tree/main | Hugging Face https://huggingface.co/agent593/Thyroid-Ultrasound-Image-Classification-Resnet50Model |
| ResNet101 | | | N/A |
| ViT-b | Torchvision https://docs.pytorch.org/vision/main/models/vision_transformer.html | HuggingFace https://huggingface.co/Wenquandan777/SARMAE/tree/main | HuggingFace https://huggingface.co/agent593/Thyroid-Ultrasound-Image-Classification-ViTModel |
| ViT-l | | | HuggingFace https://huggingface.co/sergiopaniego/fine_tuning_vit_custom_dataset_breastcancer-ultrasound-images/blob/main/model.safetensors |

**Table 8. The source of weights for each network that in the pretraining modality comparison study. Links are included for replicability.**

### *E. Best Hyperparameters – Regularization*

The highest performance achieved with each network in our regularization hyperparameter search is reported in **Table 6**, however this table does not report the values found to be optimal in each case. We report those values here in **Table 9.**

| Network | Variant | Weight Decay | DropPath |
|---|---|---|---|
| ResNet | 50 | 5% | 30% |
| | 101 | 5% | 20% |
| | 152 | 10% | 10% |
| ResNeXt | 50 | 5% | 30% |
| | 152 | 10% | 30% |
| ConvNeXt | Tiny | 20% | 30% |
| | Base | 20% | 30% |
| ViT | Small | 10% | 30% |
| | Base | 5% | 30% |
| | Large | 5% | 30% |
| SWIN | Tiny | 20% | 30% |
| | Base | 5% | 20% |
| HiViT | Small | 10% | 20% |
| | Base | 5% | 20% |

**Table 9. Values for Weight Decay and DropPath the elicited the highest performance for each network tested**